\documentclass{article} 
\usepackage[final]{colm2026_conference}

\usepackage{microtype}
\usepackage{hyperref}
\usepackage{url}
\usepackage{booktabs}
\usepackage{multirow}
\usepackage{xcolor}
\usepackage{amsmath}
\usepackage{amssymb}
\usepackage{graphicx}
\graphicspath{{figures/}}

\definecolor{darkblue}{rgb}{0, 0, 0.5}
\definecolor{darkgreen}{rgb}{0, 0.4, 0}
\definecolor{darkred}{rgb}{0.7, 0, 0}
\hypersetup{colorlinks=true, citecolor=darkblue, linkcolor=darkblue, urlcolor=darkblue}

\usepackage{lineno}

\title{Can a Lightweight Multimodal Model Estimate LLM Reasoning Performance? A Study for Compute-Optimal Document Inference}

\author{\textbf{Zishan Ahmad} \quad \textbf{Vishal Vaddina} \\
Phi Labs, Quantiphi \\
\texttt{\{zishan.ahmad, vishal.vaddina\}@quantiphi.com}
}

\newcommand{\pos}[1]{\textcolor{darkgreen}{#1}}
\newcommand{\negg}[1]{\textcolor{darkred}{#1}}

\begin{document}

\ifcolmsubmission
\linenumbers
\fi

\maketitle

\maketitle

\begin{abstract}
Modern frontier LLMs expose controllable reasoning budgets at inference time. However, allocating these budgets uniformly across all inputs is both expensive and often counterproductive, a phenomenon commonly known as the \textit{over-thinking penalty}. Frontier providers offer in-flight controls to mitigate this, such as Gemini's dynamic auto-thinking, but these mechanisms rely on the model's own self-assessment of difficulty. Recent work shows this is insufficient: 32\% of nominally cheaper model pairs cost \emph{more} in practice due to thinking-token overrun, with up to 9.7$\times$ per-query cost variance for identical prompts~\citep{chen2026reversal}. For document tasks, where difficulty is driven by visual layout, table structure, and multi-modal content, a pre-flight estimator that reads the document directly is better positioned to make this judgment than a model reasoning in-flight. Resolving this requires an \emph{external} pre-flight estimator that can predict, per sample, how well a given model will perform at each budget level \emph{before} committing to expensive inference. In this work, we explore if a lightweight multimodal model can accurately estimate the reasoning performance of a much larger LLM across variable reasoning budgets. To study this, we introduce \textbf{BudgetDoc}, the first multimodal benchmark providing explicit supervision for model-budget-performance trade-offs across three document-centric tasks. We train \textbf{DRB} (Document-Reasoning Balancer), a $\sim$1B-parameter estimator (SigLIP-2 + Qwen3-0.6B) that predicts a 7-class ordinal performance label for any (document, prompt, model, budget) configuration, achieving a weighted F1 of \textbf{0.753} on the BudgetDoc test set. We then apply DRB's estimates to optimize reasoning budgets across five frontier models and three datasets: in 9 of 15 model-dataset configurations, DRB matches or improves F1 over always-maximum-budget inference while mostly reducing cost. As a secondary case study, we probe whether the same estimator can generalize to model selection across intra- and inter-family configurations, finding encouraging preliminary results and identifying dedicated multi-provider training as an important direction for future work.
\end{abstract}

\section{Introduction}

A defining trend in large language model (LLM) development is the shift toward scaling compute at \emph{test time}~\citep{snell2024scaling,wu2024inference}. Modern frontier models, including OpenAI's GPT-5 family and Google's Gemini series, expose user-controllable reasoning budgets, enabling chains of thought~\citep{wei2022chain} of variable depth and cost. The implicit assumption is that more reasoning tokens yield better answers. In practice, this is only conditionally true.

\begin{figure}[t]
  \centering
  \includegraphics[width=\linewidth]{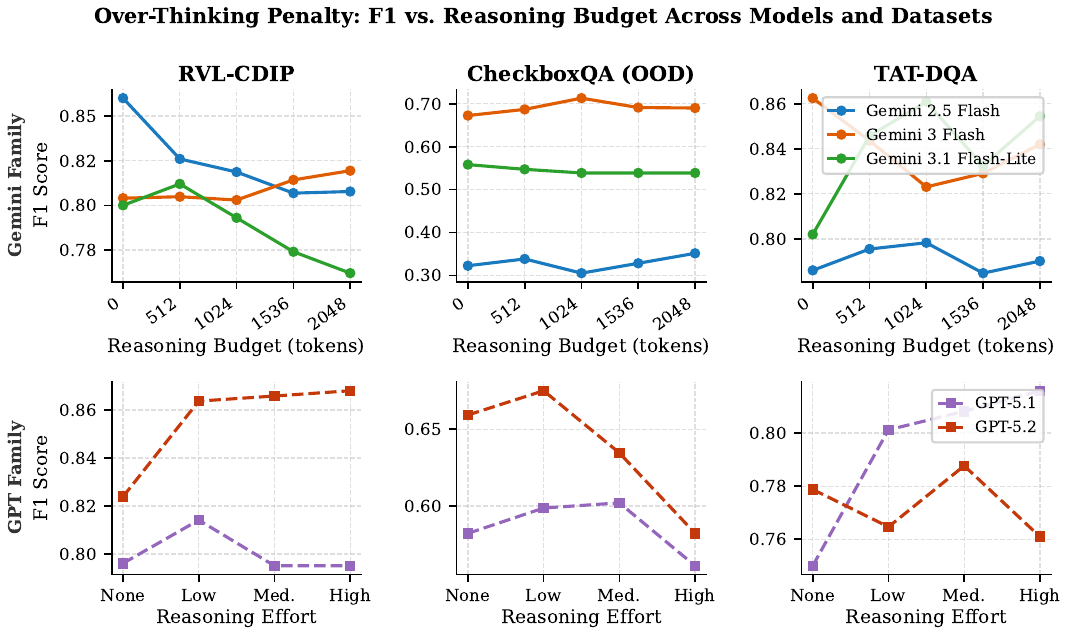}
  \caption{\textbf{The over-thinking penalty across model families and datasets.} Each line shows mean F1 as a function of reasoning budget (Gemini, top row) or reasoning effort tier (GPT, bottom row). For Gemini~2.5~Flash on RVL-CDIP, F1 \emph{decreases monotonically} from 0.86 (budget~$=0$) to 0.81 (budget~$=2048$), a 6\% penalty from over-allocation. GPT-5.2 exhibits the same pattern on CheckboxQA (0.68 at \texttt{low} $\rightarrow$ 0.58 at \texttt{high}). The optimal budget is task- and model-specific, motivating per-sample estimation rather than a fixed ceiling.}
  \label{fig:overthinking}
\end{figure}

Consider a document classification query that requires visual pattern matching rather than multi-step deduction: forcing the model through thousands of reasoning tokens does not help and may hurt, by encouraging spurious over-confident reasoning chains. We call this the \textit{over-thinking penalty}: the phenomenon where additional test-time compute leads to performance degradation or merely wastes money relative to what a lower-cost configuration would have produced. Our empirical analysis across three benchmarks and five frontier models confirms this penalty is systematic and significant (Figure~\ref{fig:overthinking}).

Providers have introduced in-flight controls to mitigate over-spending: Gemini~2.5~Flash offers a dynamic thinking mode that automatically scales its internal reasoning chain to perceived task complexity (enabled by default), while the GPT-5 family exposes a \texttt{reasoning\_effort} parameter with discrete tiers (\texttt{low}, \texttt{medium}, \texttt{high}, \texttt{xhigh}). However, these controls are fundamentally \emph{in-flight and self-assessed}: the model calibrates its own reasoning based on its own perception of difficulty, without an external estimate of its likely performance. Recent work shows this is insufficient to eliminate cost pathologies. \citet{chen2026reversal} evaluate eight frontier reasoning models and find that 32\% of nominally cheaper model-pair comparisons exhibit a \emph{pricing reversal}: the cheaper-listed model incurs higher actual cost due to uncontrolled thinking-token consumption, with per-query cost variance reaching 9.7$\times$ for identical prompts even within a single model. \citet{bai2026agents} document up to 30$\times$ within-problem token variance in agentic coding tasks, finding that models' self-predictions of their own usage achieve only weak correlation with realized consumption (Kendall $\tau \leq 0.39$). These findings establish that in-flight self-assessment is insufficient for reliable budget control, motivating an \emph{external}, pre-flight approach.

Addressing the over-thinking penalty via external estimation requires knowing, \emph{before inference}, how well a given model will perform at a given reasoning budget for a specific input. This is a non-trivial prediction problem: the performance surface over budgets varies with document complexity, query type, and model inductive biases. This raises a natural scientific question: \textbf{can a lightweight multimodal model learn to estimate the reasoning performance of a much larger LLM on a per-sample basis, enabling hard budget caps to be imposed at API-call time?}

If the answer is yes, a cheap pre-flight estimation pass could select the minimum sufficient reasoning budget for each input, achieving near-oracle performance at a fraction of the cost. The key challenge is that the estimator must be small enough that its own cost is negligible relative to the savings it enables.

To study this question rigorously, we make three contributions:

\begin{enumerate}
   \item \textbf{BudgetDoc Dataset.} We introduce the first multimodal benchmark designed specifically for supervised LLM performance estimation. BudgetDoc spans three document-centric tasks, two frontier model families, and variable reasoning budgets. By exhaustively evaluating a core set of 360 unique base samples (120 per task) across all 25 model-budget configurations, we generate a complete, dense combinatorial matrix of 9,000 labeled pairs that capture exactly how model accuracy varies with compute.
  \item \textbf{Document-Reasoning Balancer (DRB).} We train a $\sim$1B-parameter multimodal estimator composed of a SigLIP-2 vision encoder~\citep{tschannen2025siglip2}, a Qwen3-0.6B language encoder~\citep{yang2025qwen3}, a cross-page fusion transformer, and a lightweight MLP head. DRB predicts a 7-class ordinal performance label for any (document, prompt, model, budget) tuple, achieving a weighted F1 of 0.753 on the held-out BudgetDoc test set, demonstrating that small models \emph{can} learn meaningful performance estimates for much larger reasoning models.
  \item \textbf{Downstream Applications.} We demonstrate two applications of DRB's estimates: (i) \textbf{per-sample reasoning budget optimization} (primary): using DRB predictions to select the minimum budget expected to achieve near-peak performance, delivering upto 99\% cost reductions across five frontier models; and (ii) \textbf{model selection} (exploratory case study): probing whether the same estimator generalizes to selecting among models within and across provider families, finding encouraging results that motivate future dedicated work.
\end{enumerate}

Our central finding is that \textbf{a $\sim$1B-parameter multimodal model can meaningfully estimate the reasoning performance of frontier LLMs on document tasks}, with estimation quality sufficient to drive substantial cost savings through budget optimization. The over-thinking penalty is universal across modern model families: maximum reasoning budgets are routinely suboptimal in both accuracy and cost, and sample-level performance estimation is a practical lever to address this.

\section{Related Work}

\paragraph{Test-time compute scaling.}
\citet{wei2022chain} demonstrated that eliciting reasoning chains at inference time substantially improves LLM accuracy on complex tasks. Subsequent work on scaling laws for test-time compute~\citep{snell2024scaling,wu2024inference} showed that increased reasoning tokens correlate with better performance, but only for tasks of sufficient complexity. Our work identifies the complementary regime: for many inputs, over-allocated reasoning is harmful or wasteful. We frame this as an estimation problem: predicting, per sample, how performance varies with budget, rather than as a fixed-scaling problem.

\paragraph{In-flight reasoning controls and pricing reversals.}
Frontier providers have introduced mechanisms to reduce unnecessary reasoning spend. Google's Gemini 2.5 Flash uses dynamic auto-thinking (default \texttt{thinking\_budget=-1}) to scale reasoning tokens to perceived task complexity. OpenAI's GPT-5 family exposes a \texttt{reasoning\_effort} parameter with levels from \texttt{minimal} to \texttt{xhigh}. Despite these controls, \citet{chen2026reversal} show they do not eliminate cost pathologies: across 8 reasoning models and 12 tasks, 32\% of nominally cheaper model pairs cost \emph{more} in practice due to thinking-token overrun, a ``pricing reversal'' driven by the fact that thinking tokens account for 98\%+ of cost differences on single-turn tasks and exhibit up to 9.7$\times$ per-query variance for identical prompts. \citet{bai2026agents} document the same phenomenon in agentic coding tasks (up to 30$\times$ within-problem variance), showing that self-prediction by the model itself achieves only weak correlation ($\tau \leq 0.39$) with actual consumption. In our work, rather than improving in-flight self-assessment, we train an external pre-flight estimator that imposes a hard budget cap \emph{before} inference begins.

\paragraph{LLM performance prediction and evaluation.}
Predicting the output quality of large models is an emerging concern in efficient inference. Prior work has largely focused on post-hoc evaluation, self-consistency scoring~\citep{wei2022chain}, or learned reward models trained for alignment. Our work is distinct: DRB is a \emph{pre-hoc} estimator that predicts performance \emph{before} inference runs, conditioned on a configuration (model, reasoning budget) the system has not yet executed. To our knowledge, BudgetDoc is the first benchmark designed specifically for this LLM performance estimation.

\paragraph{LLM routing and model cascades.}
A related line of work addresses routing queries across multiple models to balance cost and quality~\citep{ong2024routellm}. These approaches typically route to either a strong expensive model or a weak cheap one based on query difficulty. Our primary contribution is distinct in two ways: (i) we study \emph{performance estimation} as the core task, and (ii) our main application is \emph{budget selection within a fixed model} rather than model selection. We additionally probe our estimator as a model selector in an exploratory case study, but a dedicated routing architecture is beyond the scope of this work.

\paragraph{Document understanding with VLMs.}
Document intelligence has benefited from vision-language models capable of jointly processing rendered page images and text~\citep{harley2015evaluation,zhu2021tat,turski2025uncheckedoverlookedaddressingcheckbox}. Our work builds on these benchmarks to study inference efficiency in the document understanding setting, where inputs are multi-page, visually rich, and paired with structured or open-ended queries. These properties make document tasks a particularly interesting testbed for performance estimation.

\paragraph{Backbone architectures.}
DRB draws on the success of large vision encoders (SigLIP-2~\citep{tschannen2025siglip2}) and efficient language models (Qwen3~\citep{yang2025qwen3}). The Transformer architecture~\citep{Vaswani+2017} underpins the cross-page fusion module.

\section{Methodology}

\subsection{BudgetDoc: A Benchmark for LLM Performance Estimation}
\label{subsec:budget_doc}

BudgetDoc is a supervised estimation dataset designed to answer the question: given a (document, prompt, model, reasoning budget) configuration, what performance class will the model achieve? We build it by exhaustively evaluating five frontier models from two provider families across three source benchmarks and multiple discrete reasoning budget levels, yielding a labeled dataset of configuration-outcome pairs.

\paragraph{Source benchmarks.}
\textit{(i) RVL-CDIP (Multi-class classification):} We use the Ryerson Vision Lab Complex Document Information Processing dataset~\citep{harley2015evaluation}. To increase task difficulty, we modify the single-class task by concatenating three PDFs from different categories into one multi-page document, requiring page-level multi-class prediction.
\textit{(ii) TAT-QA (Arithmetic reasoning):} The Tabular And Textual dataset~\citep{zhu2021tat} evaluates hybrid numerical reasoning over semi-structured tables and associated text in financial reports.
\textit{(iii) CheckboxQA (Form-checkbox extraction):} We use CheckboxQA~\citep{turski2025uncheckedoverlookedaddressingcheckbox} to test performance on form understanding and checkbox-state extraction. Inputs are scanned form images; outputs are structured key-value pairs derived from checkbox regions, making this a visually structured extraction task distinct from classification and arithmetic reasoning.

\paragraph{Data collection and labeling.}
To construct the benchmark pairs, we select a core set of exactly 120 unique base samples (documents and accompanying prompts) from each of the three source tasks, totaling 360 unique base contexts. We subject these samples to an exhaustive combinatorial evaluation across all five target models (three Gemini models: \texttt{gemini-2.5-flash}, \texttt{gemini-3-flash-preview}, \texttt{gemini-3.1-flash-lite-preview}; two GPT models: \texttt{gpt-5.1}, \texttt{gpt-5.2}) and five token reasoning budgets $b \in \{0, 512, 1024, 1536, 2048\}$. For the GPT family, budget levels map directly to effort tiers ($0 \!\to\! \texttt{none}$, $512 \!\to\! \texttt{low}$, $1024/1536 \!\to\! \texttt{medium}$, $2048 \!\to\! \texttt{high}/\texttt{xhigh}$). This exhaustive profiling sweeps the complete model-budget performance surface for every underlying document, yielding a dense, closed combinatorial grid of exactly 9,000 total labeled pairs ($360 \text{ base contexts} \times 5 \text{ models} \times 5 \text{ budgets}$). Prompts are paraphrased using \texttt{gemini-2.5-flash-lite} to prevent syntactic overfitting. 

Per-sample F1 scores for each configuration are discretized into seven ordinal performance classes $\mathcal{Y} \in \{0, \ldots, 6\}$ based on operational accuracy thresholds:
\begin{equation}
\small
y = \begin{cases}
0 & \text{if } f_1 < 0.5 \\
k \in \{1,\ldots,4\} & \text{if } f_1 \in [0.4 + 0.1k,\; 0.5 + 0.1k) \\
5 & \text{if } f_1 \in [0.9, 0.99) \\
6 & \text{if } f_1 \geq 0.99
\end{cases}
\end{equation}
This discretization allows the estimator to distinguish total failures (Class~0) from strong partial successes (Classes 1--5) and perfect outputs (Class~6). To prevent data leakage and ensure rigorous generalization to completely unseen text and layouts, the dataset split is strictly enforced at the unique base sample (document) level rather than via random row splitting. Out of the 360 total base contexts, 240 documents are assigned to the training set (yielding $240 \times 25 = 6,000$ training pairs), 20 documents are assigned to validation ($20 \times 25 = 500$ validation pairs), and the remaining 100 core documents are strictly held out as the test set ($100 \times 25 = 2,500$ test pairs). Because the test split preserves the complete, unbroken combinatorial grid for these 100 entirely unobserved documents, it provides a leak-free empirical ground truth to validate our out-of-sample hierarchical scanning trajectories. The 2,500-instance test split is used strictly to evaluate the estimator's raw classification performance (reporting precision, recall, and F1-score). For the downstream deployment testing, we evaluate the pipeline at the document level using the held-out base of the 100 unique test documents. Crucially, the downstream deployment pipeline requires no pre-existing configuration matrices to be present.

\subsection{DRB: A Lightweight Multimodal Performance Estimator}

The DRB model is a lightweight multimodal estimator that maps a (document, prompt, model, budget) configuration to a predicted performance class. It consists of four modality-specific encoders fused via a late-fusion MLP classification head.

\paragraph{Visual encoding.}
Document pages are encoded with a SigLIP-2-Large-patch16-512 backbone~\citep{tschannen2025siglip2} (428M parameters). For multi-page documents, page-level embeddings are processed by a 12-layer decoder-style Llama fusion transformer with hidden dimension 1152, which performs cross-page attention and mean-pooling to produce a fixed document representation $\mathbf{z}_\text{doc} \in \mathbb{R}^D$.

\paragraph{Textual encoding.}
The task prompt is encoded by Qwen3-0.6B~\citep{yang2025qwen3}, producing $\mathbf{z}_\text{prompt}$. A linear projection $\mathbf{W}_p$ aligns the textual and visual dimensions: $\mathbf{z}'_\text{prompt} = \mathbf{W}_p \mathbf{z}_\text{prompt}$.

\paragraph{Configuration conditioning.}
Learned embedding layers map the target model name $m$ and reasoning budget $b$ to continuous vectors $\mathbf{e}_\text{model}, \mathbf{e}_\text{budget} \in \mathbb{R}^D$, allowing the estimator to condition its performance prediction on the specific configuration being queried.

\paragraph{Fusion and classification.}
The four representations are concatenated and passed to a 2048-unit MLP head with GELU activation:
\begin{equation}
\mathbf{z}_\text{final} = [\mathbf{z}_\text{doc} \;\|\; \mathbf{z}'_\text{prompt} \;\|\; \mathbf{e}_\text{budget} \;\|\; \mathbf{e}_\text{model}], \qquad \hat{y} = \text{MLP}(\mathbf{z}_\text{final})
\end{equation}
We optimize with standard cross-entropy loss over the seven performance classes.

\subsection{From Estimates to Budget Decisions: Hierarchical Scanning}

At inference time, DRB's predictions are used to select a reasoning budget via hierarchical scanning. For a fixed target model, DRB queries each budget level from lowest to highest, predicting the expected performance class for each. Scanning terminates early when DRB predicts Class~6 (perfect performance), as there is no benefit to testing higher budgets. The budget with the highest predicted performance class is selected, with lower cost as a tie-breaker. In the exploratory model-selection case study (Section~\ref{sec:casestudy}), DRB scores all (model, budget) pairs jointly, though this extends the estimator beyond its training distribution.

\section{Experiments}

\subsection{Setup}

\paragraph{Models evaluated.}
We evaluate five frontier models spanning two provider families: \textit{Gemini} (\texttt{gemini-2.5-flash}, \texttt{gemini-3-flash-preview}, \texttt{gemini-3.1-flash-lite-preview}) across five token budgets $b \in \{0, 512, 1024, 1536, 2048\}$; and \textit{GPT} (\texttt{gpt-5.1}, \texttt{gpt-5.2}) across the same five budget classes mapped to effort tiers (four distinct levels, since classes 3 and 4 both map to \texttt{medium}; see Section~\ref{subsec:budget_doc}).

\paragraph{Evaluation benchmarks.}
We evaluate downstream performance on the same three task domains used to construct BudgetDoc: (i)~\textbf{RVL-CDIP} (document classification), (ii)~\textbf{TAT-DQA} (tabular QA), and (iii)~\textbf{CheckboxQA} (form-checkbox extraction). Crucially, all experimental evaluations are conducted strictly using the 100 held-out base documents quarantined in our test split, ensuring that the target models are tasked with optimizing compute for entirely unseen visual layouts and structural configurations.

\paragraph{Metrics.}
We report F1-score as the primary quality metric, total API cost per sample (USD), and latency (seconds) for baseline results. For DRB based experiments, we additionally report an \textit{adjusted cost} that factors in the estimator's infrastructure compute overhead. This overhead is calculated dynamically for each model-dataset configuration by multiplying its empirical average scanning latency by the hardware operational cost (\$0.00015 per second for a GCP T4 GPU instance, derived from standard market rates of \$0.54/hour). This total system cost ensures that all comparisons reflect the actual local resource expenditure.

\paragraph{Baselines.}
For single-model experiments, the baseline is the same model run always at \emph{maximum} reasoning budget, the natural default when no estimator is available. For the model-selection case study, the baseline is the \emph{best single model at maximum budget} per dataset, representing an oracle that already knows which model to deploy.

\subsection{How Well Does DRB Estimate LLM Performance?}
\label{sec:estimator-quality}

Before applying DRB's estimates downstream, we characterize estimation quality on the BudgetDoc test set ($n = 2500$). 
 DRB achieves a weighted F1 of \textbf{0.753}, demonstrating that a $\sim$1B-parameter model can meaningfully estimate the performance of frontier LLMs across variable reasoning budgets. Table~\ref{tab:drb-perf} (Appendix~\ref{app:drb-perf}) provides a per-class breakdown.

The most salient finding is the strong Class~6 recall of \textbf{0.932}. Class~6 represents samples where the model achieves near-perfect output regardless of budget. High recall on this class means DRB reliably detects ``easy'' samples, specifically inputs where inference will succeed without extended reasoning. This is the most actionable signal for budget optimization: if DRB is confident a sample belongs to Class~6 at a low budget, it can assign zero additional thinking tokens without sacrificing accuracy.

Estimation is hardest at the intermediate classes (F1 of 0.46--0.74 for Classes 1--5), reflecting the stochastic nature of partial reasoning successes where small differences in chain length can produce large accuracy swings. This difficulty is acceptable for the budget optimization application: even imperfect middle-class estimates allow cost reductions, as long as the system avoids misclassifying Class~6 samples as failures.

\subsection{Application I: Per-Sample Reasoning Budget Optimization}
\label{sec:budget-opt}

We now use DRB's performance estimates to dynamically select reasoning budgets per sample across the held-out test documents, replacing the naive default of always allocating the maximum budget ceiling. Table~\ref{tab:single-model} reports these out-of-sample optimization results for all five models across the three benchmark tasks.

\paragraph{DRB estimates drive consistent cost reductions.}
In 13 of the 15 model-dataset configurations, DRB-guided budget selection reduces cost relative to the maximum-budget baseline. Reductions range from 5.2\% to 99\%
, with the largest savings on GPT models where extended thinking is especially expensive. This confirms that DRB's estimates are actionable: across diverse models and tasks, they reliably identify when less reasoning is sufficient.

\paragraph{Accuracy is maintained or improved.}
In 9 of 15 configurations, DRB-guided selection matches or \emph{improves} F1 over the maximum-budget baseline, a direct consequence of the over-thinking penalty. The gains are largest for \texttt{gemini-2.5-flash} on CDIP (+5.7\%), 
 on TAT-DQA (+5.2\%) and CheckboxQA (+5.1\%). This confirms that DRB's estimation of when additional budget hurts is well-calibrated.

\paragraph{GPT cost savings are extreme.}
GPT models with extended thinking enabled are disproportionately expensive at maximum budget (c.f Table \ref{tab:single-model}, \ref{tab:gpt-effort}): \texttt{gpt-5.2} costs \$0.096/sample on CDIP and \$0.318/sample on CheckboxQA (10--60$\times$ above Gemini). DRB correctly estimates that most samples do not require this compute, selecting zero or minimal budgets and achieving 99.0\% and 98.5\% cost reductions 
with F1 changes of $\leq 3.2\%$ on CheckboxQA. This is the most compelling demonstration that estimation-guided budget selection has practical value.

\paragraph{TAT-DQA is the hardest estimation target.}
Across all models, TAT-DQA yields the weakest estimation-to-accuracy transfer: some configurations show F1 losses of up to $-5.7\%$ under DRB-guided selection. 

\begin{table}[t]
\centering
\caption{Reasoning budget optimization via DRB estimation on held-out test documents. Baseline: auto-budget for Gemini and high/xhigh for GPT models (the default without an estimator). $\Delta$F1 and Adj. $\Delta$Cost are relative to the baseline, where the adjusted cost dynamically factors in the configuration-level estimator compute overhead. $\Delta$Lat shows absolute seconds followed by percentage change. \pos{Green} = improvement, \negg{red} = degradation.}
\label{tab:single-model}
\scriptsize
\setlength{\tabcolsep}{3.5pt}
\begin{tabular}{llccccccc}
\toprule
\textbf{Dataset} & \textbf{Model} & \textbf{DRB F1} & \textbf{BL F1} & \textbf{$\Delta$F1} & \textbf{Adj.\ DRB Cost} & \textbf{BL Cost} & \textbf{Adj.\ $\Delta$Cost} & \textbf{$\Delta$Lat (s, \%)}\\
\midrule
\multirow{5}{*}{CDIP}
  & Gemini-2.5-Flash      & 0.888 & 0.840 & \pos{+5.7\%} & \$0.00169 & \$0.00427 & \pos{$-$48.0\%} & \negg{$+$3.6s ($+$27.6\%)} \\
  & Gemini-3-Flash-Preview        & 0.890 & 0.920 & \negg{$-$3.3\%} & \$0.00965 & \$0.01292 & \pos{$-$21.3\%} & \negg{$+$3.8s ($+$25.5\%)} \\
  & Gemini-3.1-Flash-Lite & 0.799 & 0.762 & \pos{+4.9\%} & \$0.00482 & \$0.00752 & \pos{$-$29.0\%} & \negg{$+$10.3s ($+$76.4\%)} \\
  & GPT-5.1               & 0.815 & 0.816 & \negg{$-$0.1\%} & \$0.00096 & \$0.01896 & \pos{$-$92.2\%} & \pos{$-$15.5s ($-$40.7\%)} \\
  & GPT-5.2               & 0.879 & 0.869 & \pos{+1.1\%} & \$0.00093 & \$0.09611 & \pos{$-$99.0\%} & \pos{$-$39.7s ($-$62.3\%)} \\
\midrule
\multirow{5}{*}{CheckboxQA}
  & Gemini-2.5-Flash      & 0.559 & 0.556 & \pos{+0.4\%} & \$0.00301 & \$0.00515 & \pos{$-$31.4\%} & \negg{$+$3.8s ($+$31.9\%)} \\
  & Gemini-3-Flash-Preview        & 0.705 & 0.670 & \pos{+5.1\%} & \$0.01746 & \$0.02126 & \pos{$-$15.4\%} & \negg{$+$3.5s ($+$19.4\%)} \\
  & Gemini-3.1-Flash-Lite & 0.668 & 0.642 & \pos{+4.0\%} & \$0.00873 & \$0.01080 & \pos{$-$14.3\%} & \negg{$+$22.0s ($+$146.9\%)} \\
  & GPT-5.1               & 0.503 & 0.478 & \pos{+5.2\%} & \$0.00274 & \$0.00202 & \negg{+35.6\%} & \negg{$+$6.5s ($+$45.5\%)} \\
  & GPT-5.2               & 0.651 & 0.673 & \negg{$-$3.2\%} & \$0.00428 & \$0.31805 & \pos{$-$98.5\%} & \pos{$-$144.3s ($-$85.4\%)} \\
\midrule
\multirow{5}{*}{TAT-DQA}
  & Gemini-2.5-Flash      & 0.514 & 0.504 & \pos{+2.1\%} & \$0.00088 & \$0.00149 & \pos{$-$5.2\%} & \pos{$-$0.9s ($-$26.4\%)} \\
  & Gemini-3-Flash-Preview        & 0.542 & 0.515 & \pos{+5.2\%} & \$0.00080 & \$0.00372 & \pos{$-$64.3\%} & \pos{$-$2.1s ($-$29.2\%)} \\
  & Gemini-3.1-Flash-Lite & 0.536 & 0.568 & \negg{$-$5.7\%} & \$0.00040 & \$0.00156 & \pos{$-$40.6\%} & \pos{$-$0.6s ($-$15.5\%)} \\
  & GPT-5.1               & 0.504 & 0.519 & \negg{$-$2.9\%} & \$0.00032 & \$0.00031 & \negg{+3.2\%} & \negg{$+$0.8s ($+$35.3\%)} \\
  & GPT-5.2               & 0.531 & 0.547 & \negg{$-$2.9\%} & \$0.00057 & \$0.06084 & \pos{$-$98.2\%} & \pos{$-$29.2s ($-$90.5\%)} \\
\bottomrule
\end{tabular}
\end{table}

\paragraph{GPT effort-level sensitivity.}
Table~\ref{tab:single-model} compares DRB against the maximum-effort GPT baseline (\texttt{high}/\texttt{xhigh}).
Table~\ref{tab:gpt-effort} (Appendix~\ref{app:gpt-effort}) extends this comparison across all four GPT effort tiers, confirming that DRB's F1 almost matches the performance at all the effort tiers. The only constant exception being TAT-DQA where the performance is marginally but consistently lower. 

\subsection{Application II: Model Selection (Case Study)}
\label{sec:casestudy}

DRB was designed and trained to estimate performance for a fixed model across budgets, not to select among models. As a secondary case study, we probe whether its learned utility surface generalizes to the more challenging task of cross-model selection: given an unobserved document input from the held-out test split and a target candidate pool, can DRB accurately identify the most resource-efficient model configuration to deploy? This question is exploratory; we do not claim DRB is a routing system, and we identify dedicated multi-provider routing architectures as important future work.

Table~\ref{tab:family} reports results for three scopes of model selection: within the Gemini family, within the GPT family, and across both families. The baseline is the oracle best single model at maximum budget (for GPT-family) or auto-budget (for Gemini-family).

\begin{table}[!t]
\centering
\caption{Exploratory case study: applying DRB estimates to cross-model selection on held-out test documents. Oracle baseline represents the best single model deployed at default auto-budget (Gemini) or maximum reasoning effort (GPT). Adjusted DRB cost dynamically incorporates the configuration-level estimator compute overhead. While cross-family selection triggers expected F1 degradation under this unobserved distribution shift, cost savings remain substantial across all scopes by routing away from expensive high-effort tiers.}
\label{tab:family}
\scriptsize
\setlength{\tabcolsep}{5pt}
\begin{tabular}{llcccccc}
\toprule
\textbf{Selection Scope} & \textbf{Dataset} & \textbf{DRB F1} & \textbf{Oracle F1} & \textbf{$\Delta$F1} & \textbf{Oracle Cost} & \textbf{Adj.\ DRB Cost} & \textbf{Adj.\ $\Delta$Cost} \\
\midrule
\multirow{3}{*}{Gemini Family}
  & CDIP        & 0.897 & 0.920 & \negg{$-$2.5\%}  & \$0.01292 & \$0.00593 & \pos{$-$54.1\%} \\
  & CheckboxQA  & 0.644 & 0.670 & \negg{$-$3.9\%}  & \$0.02126 & \$0.00513 & \pos{$-$75.9\%} \\
  & TAT-DQA     & 0.512 & 0.568 & \negg{$-$9.9\%}  & \$0.00156 & \$0.00141 & \pos{$-$9.6\%}  \\
\midrule
\multirow{3}{*}{GPT Family}
  & CDIP        & 0.781 & 0.869 & \negg{$-$10.1\%} & \$0.09611 & \$0.00121 & \pos{$-$98.7\%} \\
  & CheckboxQA  & 0.622 & 0.673 & \negg{$-$9.3\%} & \$0.31805 & \$0.00183 & \pos{$-$99.4\%} \\
  & TAT-DQA     & 0.497 & 0.545 & \negg{$-$9.5\%} & \$0.06084 & \$0.00061 & \pos{$-$99.0\%} \\
\midrule
\multirow{2}{*}{Cross-Family}
  & CDIP        & 0.861 & 0.920 & \negg{$-$6.4\%}  & \$0.01292 & \$0.00476 & \pos{$-$63.2\%} \\
  & CheckboxQA  & 0.602 & 0.676 & \negg{$-$10.9\%} & \$0.02825 & \$0.00133 & \pos{$-$95.2\%} \\
  & TAT-DQA     & 0.505 & 0.569 & \negg{$-$11.2\%} & \$0.00156 & \$0.00063 & \pos{$-$59.6\%} \\
\bottomrule
\end{tabular}
\end{table}

\paragraph{Gemini-family selection: estimation transfers well.}
Selecting among three Gemini models, the family closest to DRB's training distribution, reduces cost by 10--76\% (adjusted) with F1 losses of $-$2.5\% to $-$9.9\%. On CheckboxQA, DRB falls $-$3.9\% below oracle F1 while cutting cost by 75.9\%; on CDIP it stays within $-$2.5\% of the oracle while halving cost. These results are encouraging given that DRB was not explicitly trained for model selection.

\paragraph{GPT-family selection: large F1 loss, extreme cost savings.}
 GPT-family selection shows larger F1 degradation: $-$10.1\% on CDIP, $-$9.3\% on CheckboxQA, and $-$9.5\% on TAT-DQA relative to the oracle. 
 For GPT-family, adjusted DRB cost reduces expenditure by 98--99\% across all three datasets, because DRB systematically routes GPT inputs to low-effort configurations. The result is a stark trade-off: DRB captures nearly all available cost savings while sacrificing 9--10\% in F1.

\paragraph{Cross-family selection: the limits of estimation transfer.}
Expanding the pool to all Gemini and GPT models compounds the calibration mismatch. F1 drops by 6.4\% (CDIP) and 11.2\% (TAT-DQA).
Despite this, the cost savings remain large (59--94\% adjusted), suggesting that DRB captures enough signal to avoid expensive configurations even when fine-grained accuracy prediction drops. These results confirm that cross-family model selection requires estimators trained with explicit multi-provider supervision.

\section{Analysis}

\paragraph{Small models can estimate LLM performance.}
Figure~\ref{fig:overthinking} confirms the over-thinking penalty is systematic and universal: Gemini~2.5~Flash on RVL-CDIP loses 6\% F1 when the budget rises from 0 to 2048 tokens; GPT-5.2 loses 12\% F1 on CheckboxQA going from \texttt{low} to \texttt{high} effort. The core question of this paper, whether a lightweight multimodal model can estimate frontier LLM reasoning performance, has an affirmative answer, with important nuances. DRB achieves 0.753 weighted F1 on BudgetDoc (c.f Table \ref{tab:drb-perf}) 
 and reliably identifies the most actionable classe \textit{i.e} Class~6 (perfect output, F1 = 0.860 with recall 0.932). This is meaningful as it enables the model to predict (with high precision and recall), the tasks that are completely solvable at every budget, helping the DRB assign lower budgets with high confidence when needed.

\paragraph{What the estimation quality enables ?}
Estimation quality of 0.75 weighted F1 is sufficient to drive large practical gains because the savings mechanism is robust to prediction errors. Even when DRB mispredicts a middle class (e.g., predicts Class~4 when the true class is Class~5), the selected budget is unlikely to be catastrophically wrong. The dominant win comes from correctly identifying high-confidence Class~6 samples, which DRB does reliably, and routing them to zero or minimal reasoning budgets.

\paragraph{DRB complements, rather than replaces, in-flight reasoning controls.}
Frontier models increasingly offer built-in budget mechanisms: Gemini 2.5 Flash's auto-thinking and GPT-5's \texttt{reasoning\_effort} tiers adjust compute in-flight based on the model's self-assessment of difficulty. DRB is architecturally complementary. In-flight controls calibrate how much the model \emph{thinks}; DRB determines \emph{whether} to allocate extended thinking at all by predicting expected performance before the API call is issued. \citet{chen2026reversal} demonstrate that Gemini 3 Flash, even with dynamic auto-thinking, generated over 60{,}000 thinking tokens on select MMLU-Pro problems while GPT-5.4 needed only 25, resulting in a realized cost 35\% higher despite a lower listed price. The auto-thinking mechanism could not override the model's internally perceived need to reason at length. An external pre-flight estimator like DRB can: if DRB predicts Class~6 at budget $b=512$, the API call is issued with \texttt{thinking\_budget}$=512$, enforcing a hard cap that the model's own mechanism would not have respected. This makes external estimation a necessary complement to in-flight controls, not a redundancy.

\paragraph{When estimation is most valuable ?}
The value of estimation scales with the gap between baseline cost and optimal cost. For GPT-5.2 on CheckboxQA, this gap is enormous (\$0.318 vs.\ \$0.004 per sample); DRB's estimates are highly actionable even when imperfect. For cheap Gemini-Lite models on short documents, the absolute gap is small and the DRB inference step's latency overhead can dominate. The practical takeaway: estimation-guided budget selection is most valuable for high-cost models on complex documents.

\paragraph{Estimation fails gracefully on hard tasks.}
TAT-DQA consistently represents the hardest estimation target across all settings. For hybrid numerical reasoning over financial tables, the performance surface over budgets is irregular: small changes in chain length can produce large accuracy swings, making it difficult for any pre-hoc estimator to learn a reliable mapping. Even here, DRB achieves substantial cost savings (40--98\%) because it can still avoid extremely expensive configurations, even when fine-grained budget ranking is imperfect.

\paragraph{Estimation generalizes uniformly across document types.}
DRB achieves strong budget optimization on all three datasets (c.f Table \ref{tab:single-model}), including the visually distinctive CheckboxQA: Gemini-3-Flash-Preview improves by +5.1\% F1 at $-$15.4\% adjusted cost, and GPT-5.2 achieves 98.5\% cost savings with only $-$3.2\% F1. These results confirm that DRB's joint visual and textual representations are sufficient to capture performance variation across structurally diverse document types, ranging from dense financial tables (TAT-DQA) to scanned form images (CheckboxQA), within a single estimator trained on the BudgetDoc benchmark.


\paragraph{Early stopping reflects estimation confidence.}
In single-model settings, DRB's hierarchical scanning tests on average 1.0–3.9 of the 5 available budget levels per sample. Although the full sample-by-sample scanning trajectories are omitted from the primary tables for brevity, a comprehensive analysis of our runtime execution logs reveals highly efficient search dynamics. On CDIP, all Gemini models achieve a 100\% early-stop rate, where the estimator is confident enough on the very first budget level to immediately terminate the scan and bypass higher tiers. This high-confidence conservative behavior holds globally: across the remaining dataset-model configurations, ensuring that the estimator's operational overhead remains modest relative to the downstream API savings it enables.

\section{Conclusion}

We asked whether a lightweight multimodal model can estimate the reasoning performance of frontier LLMs on document tasks. The answer is yes: DRB, a $\sim$1B-parameter estimator, achieves 0.753 weighted F1 on the BudgetDoc benchmark 
 and, crucially, 0.932 recall on perfect-performance class. These estimates are actionable: applying them for per-sample reasoning budget selection delivers up to 99\% cost reductions across five frontier models and three datasets, while maintaining or improving F1 in 9 of 15 configurations. The over-thinking penalty, where maximum reasoning budgets produce worse accuracy at higher cost, is systematic and universal across modern model families.

We additionally showed that DRB's learned performance surface partially transfers to model selection (an application it was not designed for), with moderate F1 losses ($-$2.5\% to $-$9.9\%) and strong cost savings (10--76\%) for intra-family Gemini selection, and degrading F1 accuracy as the scope widens to GPT and cross-family settings (though cost savings remain large at 59--99\% due to avoidance of expensive high-effort configurations). We identify multi-provider performance estimation as an important direction for future work, alongside dedicated routing architectures trained with explicit cross-provider supervision.

Our results also establish that external pre-flight estimation is necessary even as frontier providers offer increasingly sophisticated in-flight reasoning controls. Built-in dynamic thinking modes (Gemini auto-thinking, GPT reasoning\_effort tiers) adjust compute based on the model's self-assessment, but \citet{chen2026reversal} show this does not prevent pricing reversals: 32\% of nominally cheaper models still cost more in practice due to thinking-token overrun. \citet{bai2026agents} confirm that models systematically underestimate their own token consumption. An external estimator like DRB provides a hard pre-flight cap that in-flight mechanisms cannot. 
BudgetDoc and the DRB baseline are released to facilitate future research into sample-level LLM performance estimation and compute-optimal inference.

\section*{Limitations}

While our framework demonstrates robust performance across both Gemini and GPT model families, several limitations govern its deployment. First, expanding or adapting this architecture to entirely new model providers or unobserved model architectures introduces a data-gathering requirement; a small calibration sample of configuration-outcome pairs must be collected through a standard evaluation process to fine-tune the estimator on the new target domain. Second, our evaluation is strictly restricted to document-centric tasks where visual layout and structural complexity dictate reasoning success. Extending pre-flight performance estimation to non-visual, text-dominant applications such as general-purpose QA, code generation, or open-ended creative tasks remains an open direction for future work. Finally, the hierarchical budget scanning process could introduce an operational latency overhead depending on the task. This computational overhead dictates that our framework is highly favorable for token-heavy, high-cost frontier model inference, but less suitable for ultra-low-latency or real-time application pipelines.

\section*{Ethics Statement}
This work uses publicly available benchmarks (RVL-CDIP, TAT-QA, CheckboxQA) and frontier model APIs accessed under standard commercial terms. No personal data was collected or used. By enabling estimation-guided budget selection, our work reduces unnecessary reasoning token expenditure, which has a positive environmental impact.

\bibliography{colm2026_conference}
\bibliographystyle{colm2026_conference}

\appendix

\section{DRB Per-Class Estimation Performance on BudgetDoc}
\label{app:drb-perf}

Table~\ref{tab:drb-perf} provides the full per-class breakdown of DRB's estimation performance on the BudgetDoc test set ($n = 2500$). Classes 0--6 correspond to F1-score bins defined in Section~3.1, where Class~0 represents total task failure and Class~6 represents perfect output.

\begin{table}[h]
\centering
\scriptsize
\caption{DRB (SigLIP-2 + Qwen3-0.6B) per-class estimation performance on the BudgetDoc test set. Class~6 (perfect performance) has the highest F1, driven by strong recall, meaning DRB reliably identifies samples where any budget will succeed.}
\label{tab:drb-perf}
\begin{tabular}{lcccc}
\toprule
\textbf{Class (F1 bin)} & \textbf{Precision} & \textbf{Recall} & \textbf{F1} & \textbf{Support} \\
\midrule
0 \quad ($f_1 < 0.5$)              & 0.623 & 0.491 & 0.549 & 448  \\
1 \quad ($f_1 \in [0.5, 0.6)$)    & 0.646 & 0.361 & 0.463 & 87   \\
2 \quad ($f_1 \in [0.6, 0.7)$)    & 0.716 & 0.671 & 0.693 & 143  \\
3 \quad ($f_1 \in [0.7, 0.8)$)    & 0.744 & 0.569 & 0.645 & 174  \\
4 \quad ($f_1 \in [0.8, 0.9)$)    & 0.849 & 0.656 & 0.740 & 154  \\
5 \quad ($f_1 \in [0.9, 0.99)$)   & 0.781 & 0.653 & 0.711 & 98   \\
6 \quad ($f_1 \geq 0.99$)          & 0.798 & 0.932 & 0.860 & 1396 \\
\midrule
Weighted Avg                        & 0.756 & 0.765 & 0.753 & 2500 \\
\bottomrule
\end{tabular}
\end{table}

The strong Class~6 recall (0.932) is the most consequential metric for the budget optimization application: it governs how reliably DRB can assign minimal reasoning budgets to samples where any budget would yield correct output, without sacrificing accuracy. Low Class~1 recall (0.361) reflects the inherent difficulty of distinguishing near-total failures from total failures, a fine-grained distinction that matters less in practice, since both map to low-cost budget assignments.

\section{Implementation Details}
\label{app:impl}

The DRB estimator uses a \texttt{SigLIP-2-large-patch16-512} vision backbone and a \texttt{Qwen3-0.6B} prompt encoder. Training was conducted for 20 epochs on an NVIDIA A100 (80GB) using the AdamW optimizer with learning rate $5 \times 10^{-5}$ and batch size 4. The 12-layer fusion transformer uses hidden dimension 1152.

\paragraph{DRB compute overhead.} The DRB estimator is deployed on GCP \texttt{n1-standard-4} with one NVIDIA T4 GPU (us-central1 on-demand pricing: \$0.19/hr VM + \$0.35/hr GPU = \$0.54/hr total, reducing to \$0.00015 per second of compute). Because our hierarchical budget scanning uses an early-stopping mechanism, the execution latency scales dynamically based on the complexity of the document-query pair. To ensure a rigorous financial comparison, the adjusted system cost ($\text{Adj.\ DRB Cost}$) reported in paper across tables is computed dynamically at the configuration level:
\begin{equation*}
  \text{Adj.\ DRB Cost} = \text{API Cost}_{\text{RB}} + (\Delta t_{\text{config}} \times \$0.00015)
\end{equation*}
where $\Delta t_{\text{config}}$ represents the empirical mean scanning latency for that specific model-dataset group. This approach guarantees that the local computational burden of running the pre-flight estimator is fully captured and charged against the downstream API savings.

\section{GPT Effort-Level Sensitivity}
\label{app:gpt-effort}

Table~\ref{tab:gpt-effort} extends the comparison from Table~\ref{tab:single-model} to all four GPT reasoning effort tiers (\texttt{none}, \texttt{low}, \texttt{medium}, \texttt{high}/\texttt{xhigh}). 

\begin{table}[h]
\centering
\caption{DRB versus GPT models across all four reasoning effort tiers. Adj.\ $\Delta$Cost factors in the dynamic configuration-level compute overhead. \pos{Green} = improvement, \negg{red} = degradation.}
\label{tab:gpt-effort}
\scriptsize
\setlength{\tabcolsep}{3.5pt}
\begin{tabular}{llccccccc}
\toprule
\textbf{Dataset} & \textbf{Model (Effort)} & \textbf{DRB F1} & \textbf{BL F1} & \textbf{$\Delta$F1} & \textbf{Adj.\ DRB Cost} & \textbf{BL Cost} & \textbf{Adj.\ $\Delta$Cost} \\
\midrule
\multirow{8}{*}{CDIP}
  & GPT-5.1 (Low)     & 0.815 & 0.814 & $\pm$0.0\%     & \$0.00096 & \$0.01814 & \pos{$-$91.8\%} \\
  & GPT-5.1 (Medium)  & 0.815 & 0.813 & \pos{$+$0.2\%} & \$0.00096 & \$0.02148 & \pos{$-$93.1\%} \\
  & GPT-5.1 (High)    & 0.815 & 0.816 & \negg{$-$0.1\%} & \$0.00096 & \$0.02900 & \pos{$-$94.9\%} \\
  & GPT-5.2 (Low)     & 0.879 & 0.873 & \pos{$+$0.7\%} & \$0.00093 & \$0.05738 & \pos{$-$97.5\%} \\
  & GPT-5.2 (Medium)  & 0.879 & 0.876 & \pos{$+$0.3\%} & \$0.00093 & \$0.05851 & \pos{$-$97.5\%} \\
  & GPT-5.2 (High)    & 0.879 & 0.869 & \pos{$+$1.1\%} & \$0.00093 & \$0.06073 & \pos{$-$97.6\%} \\
\midrule
\multirow{8}{*}{CheckboxQA}
  & GPT-5.1 (Low)     & 0.503 & 0.491 & \pos{$+$2.4\%} & \$0.00274 & \$0.02418 & \pos{$-$86.5\%} \\
  & GPT-5.1 (Medium)  & 0.503 & 0.488 & \pos{$+$3.1\%} & \$0.00274 & \$0.03014 & \pos{$-$89.2\%} \\
  & GPT-5.1 (High)    & 0.503 & 0.478 & \pos{$+$5.2\%} & \$0.00274 & \$0.05040 & \pos{$-$93.5\%} \\
  & GPT-5.2 (Low)     & 0.651 & 0.663 & \negg{$-$1.8\%} & \$0.00428 & \$0.07439 & \pos{$-$93.5\%} \\
  & GPT-5.2 (Medium)  & 0.651 & 0.658 & \negg{$-$1.1\%} & \$0.00428 & \$0.07751 & \pos{$-$93.8\%} \\
  & GPT-5.2 (High)    & 0.651 & 0.673 & \negg{$-$3.2\%} & \$0.00428 & \$0.12162 & \pos{$-$96.0\%} \\
\midrule
\multirow{8}{*}{TAT-DQA}
  & GPT-5.1 (Low)     & 0.504 & 0.513 & \negg{$-$1.8\%} & \$0.00032 & \$0.00210 & \pos{$-$59.8\%} \\
  & GPT-5.1 (Medium)  & 0.504 & 0.508 & \negg{$-$0.8\%} & \$0.00032 & \$0.00276 & \pos{$-$69.4\%} \\
  & GPT-5.1 (High)    & 0.504 & 0.519 & \negg{$-$2.9\%} & \$0.00032 & \$0.00547 & \pos{$-$84.5\%} \\
  & GPT-5.2 (Low)     & 0.531 & 0.540 & \negg{$-$1.7\%} & \$0.00057 & \$0.00511 & \pos{$-$78.6\%} \\
  & GPT-5.2 (Medium)  & 0.531 & 0.535 & \negg{$-$0.7\%} & \$0.00057 & \$0.00605 & \pos{$-$81.9\%} \\
  & GPT-5.2 (High)    & 0.531 & 0.547 & \negg{$-$2.9\%} & \$0.00057 & \$0.00700 & \pos{$-$84.4\%} \\
\bottomrule
\end{tabular}
\end{table}

\end{document}